# Co-occurrence of medical conditions: Exposing patterns through probabilistic topic modeling of snomed codes

Moumita Bhattacharya[a,*], Claudine Jurkovitz[b], Hagit Shatkay[a,c,d]

[a] Computational Biomedicine Lab, Computer and Information Sciences, University of Delaware, Newark, DE, USA
[b] Value Institute, Christiana Care Health System, Newark, DE, USA
[c] Center for Bioinformatics and Computational Biology, Delaware Biotechnology Inst, University of Delaware, DE, USA
[d] School of Computing, Queen's University, Kingston, ON K7L 3N6, Canada

A B S T R A C T

Patients associated with multiple co-occurring health conditions often face aggravated complications and less favorable outcomes. Co-occurring conditions are especially prevalent among individuals suffering from kidney disease, an increasingly widespread condition affecting *13%* of the general population in the US. This study aims to identify and characterize patterns of co-occurring medical conditions in patients employing a probabilistic framework. Specifically, we apply topic modeling in a non-traditional way to find associations across SNOMED-CT codes assigned and recorded in the EHRs of > *13,000* patients diagnosed with kidney disease. Unlike most prior work on topic modeling, we apply the method to codes rather than to natural language. Moreover, we quantitatively evaluate the topics, assessing their tightness and distinctiveness, and also assess the medical validity of our results. Our experiments show that each topic is succinctly characterized by a few highly probable and unique disease codes, indicating that the topics are tight. Furthermore, inter-topic distance between each pair of topics is typically high, illustrating distinctiveness. Last, most coded conditions grouped together within a topic, are indeed reported to co-occur in the medical literature. Notably, our results uncover a few indirect associations among conditions that have *hitherto not been reported* as correlated in the medical literature.

## 1. Introduction

Patients associated with multiple co-occurring health conditions often face aggravated complications and less favorable treatment outcomes. The Center for Disease Control and Prevention reports that in the US alone, one in four individuals suffers from multiple health conditions, while the rate is three times higher among individuals *65* or older [1]. Co-occurring conditions are especially prevalent among individuals diagnosed with *kidney disease*, an increasingly widespread condition affecting *13%* of the general population and *18%* of hospitalized patients in the US [2]. Kidney disease is associated with a large number of complications, including cardiovascular disease, metabolic bone disease and diabetes [3]. Identifying conditions that tend to co-occur in the context of kidney disease, followed by evidence based interventions, can slow or prevent a patient's progression to advanced stages of the condition. In this study, we thus aim to identify co-occurrence patterns of medical conditions among individuals who have kidney disease.

We analyzed, in collaboration with physicians and researchers from Christiana Care Health System, *Electronic Health Records* (*EHRs*) of > *13,000* patients, gathered over a period of eight years, from primary care and specialty practices across Delaware associated with Christiana Care, the largest health-system in Delaware. We included patients' records in our dataset, if at any time during follow-up, there was an indication of a decline in kidney function, determined by a lower than normal ($< 60\,\text{mL/min}/1.73\,\text{m}^2$), estimated *Glomerular filtration rate* (*GFR*), a common marker of kidney function. Each record includes attributes such as vital signs, demographics and SNOMED codes for diagnosed conditions, recorded during multiple *office visits* and *hospital stays*.

We focus our analysis on the diagnosed conditions attribute in the dataset, represented through the healthcare terminology of *SNOMED-CT codes* [4]. SNOMED is specifically designed to capture detailed information during clinical care, enabling clinicians to choose appropriate conditions from a predefined fine grained list, and is thus a *structured non-text* representation of patients' diagnoses. Ours is the first study that utilizes SNOMED codes for identifying patterns of co-occurring conditions. The large number of patients, the wide timespan in our EHRs and the use of SNOMED codes to represent diagnosed conditions, yield a large scale dataset which supports identifying patterns

* Corresponding author.
*E-mail addresses:* moumitab@udel.edu (M. Bhattacharya), CJurkovitz@christianacare.org (C. Jurkovitz), shatkay@udel.edu (H. Shatkay).

of co-occurring conditions.

Specifically, we adopt a data driven approach to identify many-to-many associations among a broad group of medical conditions associated with patients who have kidney disease. We apply a probabilistic topic modeling method, *Latent Dirichlet Allocation* (*LDA*) to SNOMED codes [5]. Topic modeling has primarily been used for identifying thematic structures (*topics*) in unstructured text data. In contrast, we apply topic modeling over structured non-text SNOMED codes to automatically organize diagnoses associated with patients' EHRs into groups of correlated coded conditions, where each group represents a topic, formally defined as a probability distribution over codes. In our study, a *patient file*, which includes all coded conditions with which the patient has been diagnosed, is viewed as a document, and each code is treated as a word. We hypothesize that a set of coded conditions, tending to co-occur in patients, also demonstrate a high probability to be associated with a specific topic. In preliminary studies (not shown), we employed simpler unsupervised methods such as *K*-means and cosine-similarity-based *clustering* to identify co-occurrence patterns of medical conditions, but unlike topic model those have not revealed meaningful, clinically relevant associations.

Most previous studies, aiming to identify associations among diseases have focused on exposing associations among a few pre-defined specific conditions [6–13]. For instance, Farran et al. targeted association between diabetes and hypertension [12], while a more recent work by Chen et al. explored association between colorectal cancer and obesity [13]. In contrast, ours is the first study aiming to find many-to-many associations among a broad group of conditions, doing so within a specific disease context, namely, kidney disease. Our results thus show highly informative groups of meaningful connections (i.e. *informative topics*), which manifest themselves within the context of the disease.

A handful of earlier studies explored association among a broad group of conditions [14–16] utilizing textual data present in patient records. In contrast, we utilize *non-text* SNOMED codes that are designed to unambiguously record diagnosed conditions during clinical care. Hence, using SNOMED as the basis for the analysis can directly expose connections among conditions.

Topic modeling has been broadly used for text analysis [5] and image processing [17] applications. Recently, topic models have also been applied in the biomedical domain [18–23], mostly applying the method to text data (e.g., physicians' notes, biomedical literature). Only a few studies have applied topic models to non-text data as we do here [24–27]. Most recently, a study by Chen et al. [27] used topic modeling for predicting clinical order-sets from inpatient hospitalization records, with reported performance of *47%* precision and *24%* recall. In contrast to our study, which aims to expose patterns of co-occurring medical conditions while rigorously assessing topic quality, Chen et al. aimed to predict clinical order-sets using topic modeling as tools without evaluating the obtained topics themselves.

The work most related to ours, a study by Li et al. [26], utilizes topic modeling to cluster patient diagnosis-groups, represented by ICD-9 codes, for identifying comorbidity. Notably, ICD-9 codes are defined at a coarser granularity level than SNOMED codes and thus capture less detailed information during clinical care, than SNOMED codes [28]. Moreover, the dataset used in that work was relatively small, comprising only *4644* patients. Most importantly, unlike our study, the study by Li et al. does not ensure that topics indeed capture clinically-relevant associations; nor evaluates the topics quantitatively.

Preliminary steps of this study were discussed in our earlier extended abstract [29]. Here we present additional experiments and results and extended analysis. Specifically, we conduct a thorough quantitative evaluation of the performance of our model, and an assessment of the medical validity of our results. Moreover, we include results obtained over an additional dataset consisting of many thousands of hospitalization records, which were not available for the preliminary study.

We assess the performance of our method *qualitatively* as well as *quantitatively*. For qualitative evaluation, we assess the clinical validity of our results by examining whether the conditions that show high probability to be associated within a topic are known to co-occur according to the medical literature. When applying an unsupervised method such as topic modeling to data, the uncovered groups (topics) may not always carry a useful cohesive semantics. Hence, it is important to quantitatively assess whether the topics are indeed informative. Unlike most other studies on topic modeling, we quantitatively evaluate the topics obtained from our model, by assessing their tightness and distinctiveness.

*Tightness* signifies that the codes characterizing the topic are highly correlated with each other and thus capture strong non-arbitrary associations. *Distinctiveness* across topics indicates that codes that are highly probable to be grouped under a specific topic, are unlikely to be associated with other topics. Furthermore, to verify that the topics are informative, we measure the entropy of each topic, and show entropy that is significantly lower than that of the uniform distribution (i.e. high information contents).

We applied our method to two datasets collected from the patients, one comprising office visit records and the other hospitalization records. In both cases, we obtained informative, distinct and tight topics that align well with known co-occurrences among conditions cited in the medical literature. We note that similar disease themes emerge in topics inferred from each of the two sets, but also that the conditions grouped together under topics obtained from the hospitalization dataset typically indicate higher severity than those grouped under topics inferred from office visits. The difference is expected, as patients who are hospitalized often manifest more severe conditions (such as *congestive heart failure* and *renal failure*), than patients who are not.

Our method's ability to generate meaningful topics from both datasets, where one comprises more common conditions (office visits) and the other more severe ones (hospitalization cases), demonstrates its effectiveness in reliably exposing co-occurring conditions. Notably, our results uncover a few indirect associations among conditions that have hitherto gone unreported, suggesting that topic modeling over codes can expose yet unnoticed associations.

## 2. Material and methods

### 2.1. Datasets

Our datasets consist of information gathered from office visit records of *13,111* patients in Delaware, who have shown evidence of decreased kidney function during these visits, and from the hospitalization records of a subset of *9,530* patients who had at least one hospital visit. The records included in the office visit set and those included in the hospitalization set do not overlap. Each record contains attributes such as age, ethnicity, and diagnosed conditions, collected between August 2007 and July 2015. As noted, we focus solely on the diagnosed conditions attribute, represented via SNOMED codes.

We extracted the office visit records from the Christiana Care Health System outpatient EHR (Centricity™, GE) and the hospitalization records from the Christiana Care inpatient EHR (PowerChart®, Cerner Millennium). Patients' identifiers were removed from records in both the office visit and hospitalization sets. The records were then transformed and standardized into a common data model, the *Observational Medical Outcomes Partnership* (*OMOP*) [30]. SNOMED is one of the standardized vocabularies available in the OMOP model. This project was approved by the Christiana Care IRB with a waiver of consent according to 45CFR46.116d.

Table 1 summarizes key characteristics of the office visit and the hospitalization datasets. As shown in the table, the patients' age is quite high; the mean age is *70* ($\sigma = 12$) for the office-visit set and *67* ($\sigma = 14$) for the hospitalization set. The vast majority of patients (*94%*) are above *50*, *66%* above *65,* while less than *1%* are below *35*. Fig. 1 shows descriptions of the twenty most frequent codes appearing in each record set, in decreasing order of occurrence frequency.

**Table 1**
Key characteristics of the office visit and hospitalization datasets. The leftmost column lists the characteristics; the middle column shows the corresponding values in the office visit set, while the rightmost column shows the values in the hospitalization set.

| | **Office visit set** | **Hospitalization set** |
|---|---|---|
| *Number of Patients* | 13,111 | 9,528 |
| *Age Range ($25^{th}$ –$75^{th}$ Percentile)* | 60–80 | 60–80 |
| *Mean Age* ($\sigma$) | 70 (*12*) | 67 (*14*) |
| *% Female* | 60% | 61% |
| *% Male* | 40% | 39% |
| *Avg. Number of Visits per Patient* | 17 | 5 |

To create a data-matrix that can be processed via topic modeling we make several data organization and preprocessing choices. Traditionally in topic modeling, a natural language word is the basic data unit, while the set of unique words is referred to as the vocabulary. In contrast, in our study, we use coded conditions, represented as SNOMED codes, rather than words, such that the vocabulary is the set of unique codes. To determine the number of unique codes included in our vocabulary, denoted *V*, we conducted multiple experiments while varying the vocabulary size, considering all codes occurring in the datasets and reduced code sets accounting for *90%* and for *80%* of the cumulative frequency in each dataset. As further discussed in Section 3, the larger of these vocabularies do not perform as well. We thus focus here on the vocabulary comprising the set of codes accounting for *80%* of the cumulative frequency. Of the *4,000* codes in the office visit set, just *180* account for *80%* of the cumulative frequency (see Fig. 2). To avoid sparsity in the data-matrix, we limit our vocabulary to these *180* frequent codes ($V = 180$). Similarly, when working with the hospitalization set, we limit our vocabulary to *250* of the *4,000* codes ($V = 250$), which accounts again for *80%* of the cumulative frequency.

After defining the vocabulary, we preprocess the original patient files to represent each patient in a *bag-of-codes* format (analogous to the bag-of-words representation of documents) for applying topic modeling. Bag-of-codes accounts for the number of times a code appears in a patient file. Recurring conditions that persistently recur multiple times in the file (e.g. *chronic conditions*), and often indicate a higher level of severity, thus incur a higher weight (counts). In contrast, conditions occurring only sporadically in the patient's history (e.g. occasional *cough* or *headache*) receive a lower weight, and as such play a lesser role in the topic profiles. We also experimented with two additional common representations, namely, *binary* (*0/1*) and *tf-idf* (*term frequency, inverse document frequency*), which proved less effective as we further discuss in Section 3.

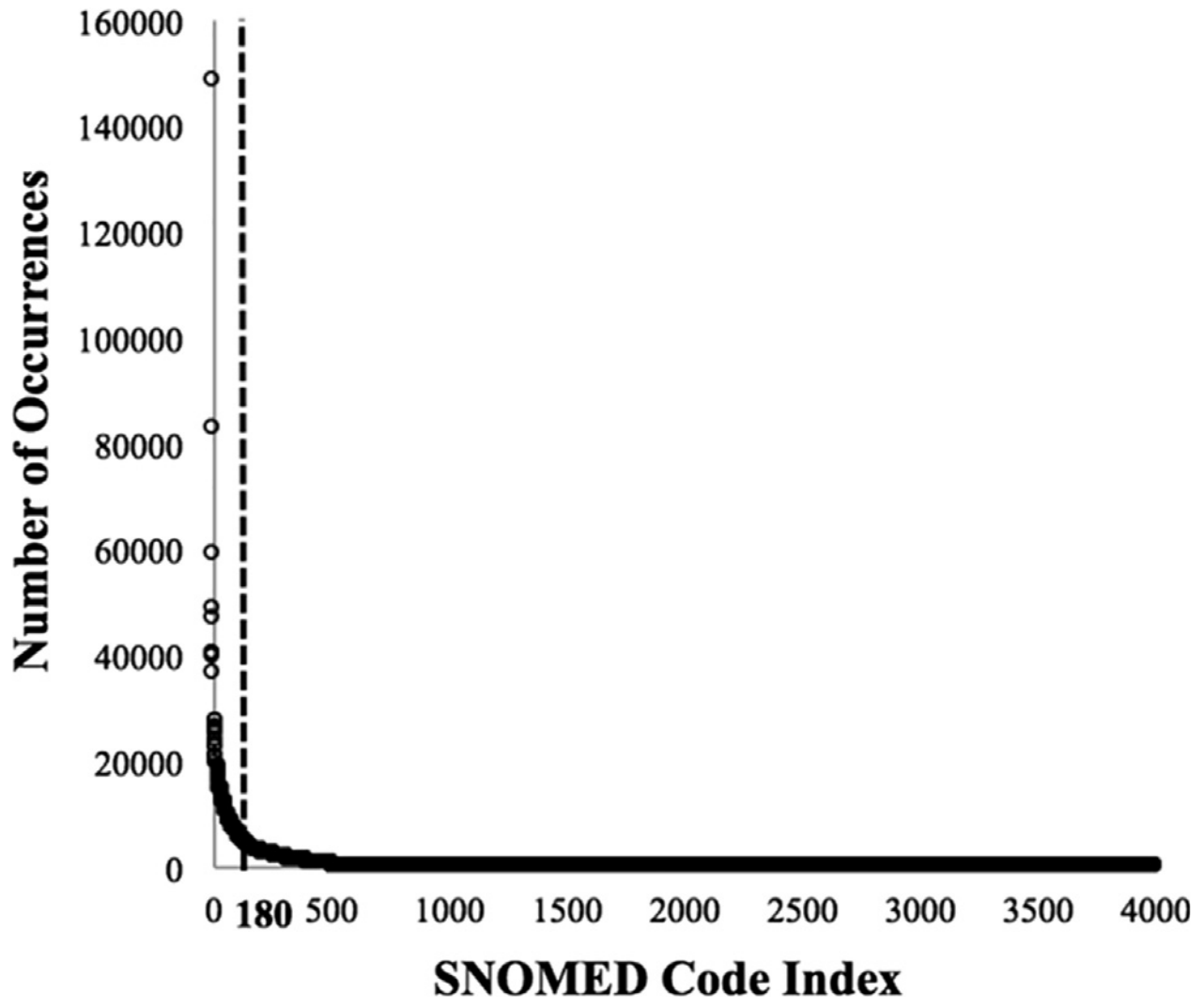

**Fig. 2.** Number of occurrences for each of the 4,000 SNOMED codes in the office visit set. The *y*-axis shows the number of times each SNOMED code occurs in the set; the *x*-axis corresponds to indices representing the 4,000 SNOMED codes, sorted in decreasing order of number of occurrences.

Each patient file, $F_i$ ($1 \le i \le M$), is converted to a vector of codes, $F_i = \langle c_1^i,...,c_{N_i}^i \rangle$, where *M* denotes the number of patients in the dataset and $N_i$ is the total number of code occurrences in the $i^{th}$ patient file. Each code, $c_j^i$, in the vector is one of the *V SNOMED codes* in our vocabulary, viewed as a value taken by a respective random variable, $C_j$ ($1 \le j \le N_i$), denoting the code value occurring in the $j^{th}$ position of the $i^{th}$ patient file. We note that any of the $V$ codes in our vocabulary can appear at any position in a patient file.

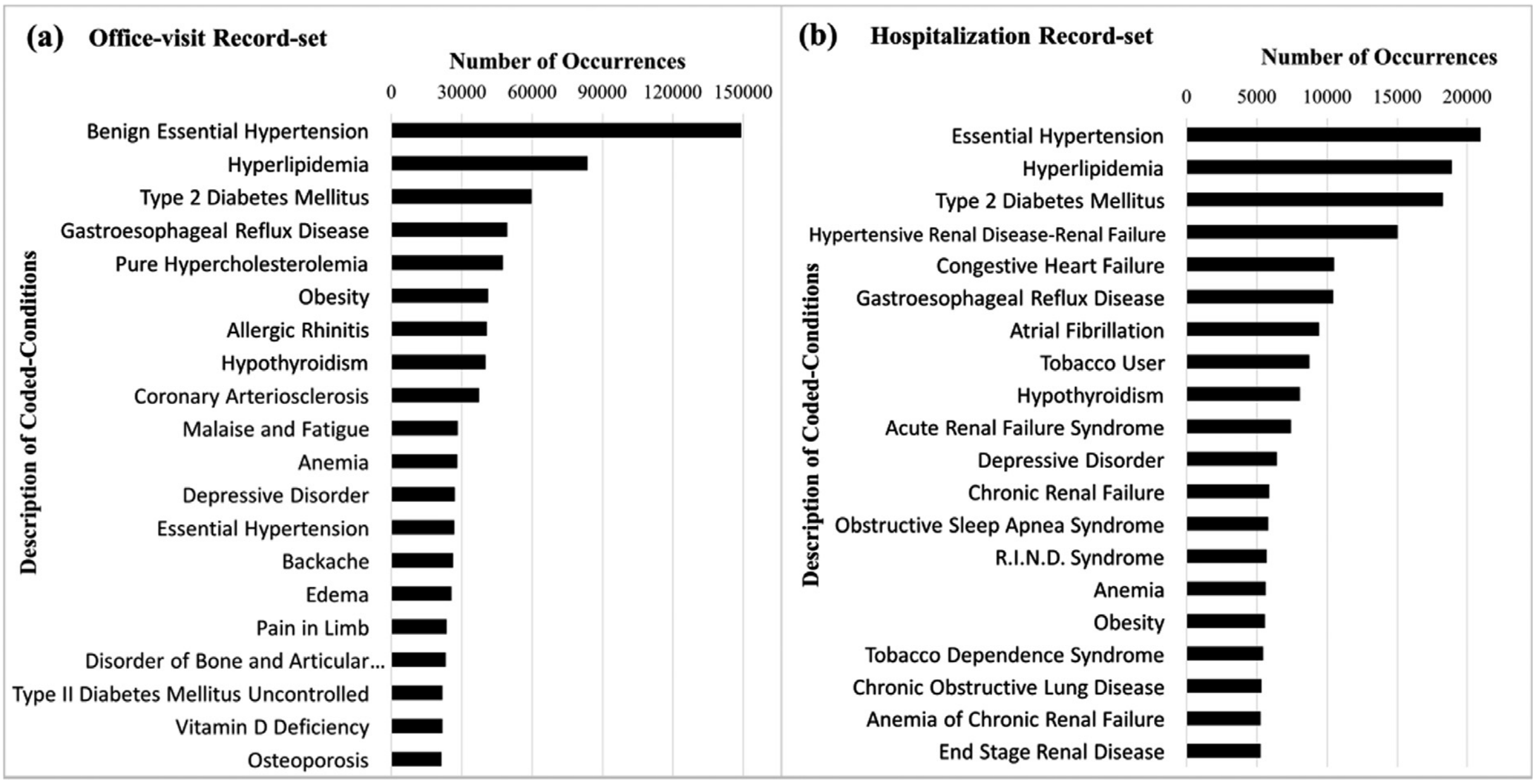

**Fig. 1.** Number of occurrences of the twenty most frequent conditions in the *office visit* record set (*a*) and in the *hospitalization* record set (*b*). Each bar represents the number of occurrences of each condition in the respective record set. For clarity, we show the descriptive name associated with each SNOMED code rather than the code itself.

We process all patient files, $F_i$, ($1 \le i \le M$), to create a data matrix where each row represents a patient, and each column represents a SNOMED code included in the vocabulary, such that each cell $\langle p, c \rangle$ holds the number of times a patient $p$ was diagnosed with condition $c$. Each patient in the record set is thus associated with a *V-dimensional* vector, where $V$ denotes the number of unique codes in our vocabulary and each vector entry corresponds to the occurrence frequency of a condition within the patient's file. We refer to each such vector as a *patient-conditions record,* and to the collection of all vectors as the *patient-conditions corpus*. The latter consists of a $13{,}111 \times 180$ matrix for the office visits dataset, and a $9{,}528 \times 250$ matrix for the hospitalization dataset.

Studies analyzing clinical data often take into account age as a possible confounding factor affecting disease severity and number of hospitalizations or office visits, and consider stratifying the population by age. In this study we applied similar consideration, and inspected the variation in the total number of code occurrences within patient records as a function of age (data not shown). We have found that the the total number of times coded-conditions are assigned to each patient do not vary significantly with age. This finding is not surprising, because as noted above, the patients in our dataset are predominantly ($>94\%$) older than *50*, and are all (even the younger ones) at moderate-to-severe stages of the disease. As such, physician visits are frequent and code distributions do not significantly vary by age throughout the population; we thus apply topic-modeling to the population as a whole without any further stratification.

Table 2 shows examples of five patient-conditions records, where each row corresponds to a patient's record and each column corresponds to a condition. Due to limited space, only eight conditions are shown. Each cell lists the number of times a patient has been diagnosed with the condition.

### 2.2. Latent Dirichlet Allocation (LDA)

We employ LDA to model patient records as though they were generated by sampling from a mixture of $K$ underlying topics, where a topic is a multinomial distribution over all coded conditions in our vocabulary [5]. The generative process for each patient file consists of the following steps:

First, a multinomial distribution over $V$ codes for the $t^{th}$ topic, denoted $\Phi_t$ ($1 \le t \le K$), is obtained by sampling from a Dirichlet distribution with parameter $\alpha$; $\Phi_t$ represents the conditional probability of a code to occur in the $t^{th}$ topic. Next, for each patient file, $F_i$, a multinomial distribution over $K$ topics, denoted $\theta_i$, is sampled from a Dirichlet distribution with parameter $\beta$; $\theta_i$ represents the conditional probability of the file to be associated with each of the $K$ topics. Subsequently, for each code-position, $j$, in the file, $F_i$: (*1*) A topic is drawn by sampling from $\theta_i$; the selected topic at position $j$ in $F_i$, is denoted $z_j^i \in \{1,\ldots,K\}$; (*2*) Given the topic $z_j^i$ a code $c_j^i$ is drawn by sampling the topic-code distribution, $\Phi_{z_j^i}$.

The model parameters are learnt iteratively for different values of $K$, where $K$ ranges from *5* to *100*, and the data log-likelihood is calculated for each value of $K$. To determine the optimal number of topics, we identify the $K$ value that maximizes the data log-likelihood, which is defined as: $\sum_{i=1}^{M} log \int \left\{ \sum_z \left[ \prod_{j=1}^{N_i} Pr(c_j^i | z_j^i, \Phi_{z_j^i}) Pr(z_j^i | \theta_i) \right] \right\} Pr(\theta_i | \beta) d\theta_i$, (where $M$ denotes the number of patients in the corpus and $N_i$ denotes the total number of code occurrences in the $i^{th}$ patient file, as defined earlier. See Hornik et al. [31] for details.)

To learn the model parameters based on our data and obtain the data log-likelihood, we employ the *R topicmodels* library [32], which uses Gibbs sampling. We use the default parameter values, set in the *topicmodels* library, for the parameter $\beta$ (*0.1*) and for the initial value of the parameter $\alpha$ (*50/M*) [31].

### 2.3. Jensen-Shannon divergence (JSD)

*JSD* is a symmetric measure of similarity between two probability distributions [33]. Let $\overrightarrow{X} = \langle x_1,\ldots,x_N \rangle$ and $\overrightarrow{Y} = \langle y_1,\ldots,y_N \rangle$ represent two *N-dimensional* multinomial distributions. The *JSD* between $\overrightarrow{X}$ and $\overrightarrow{Y}$ is defined as: $JSD(\overrightarrow{X} || \overrightarrow{Y}) = \frac{1}{2} \sum_{i=1}^{N} x_i log\left(\frac{x_i}{m_i}\right) + \frac{1}{2} \sum_{i=1}^{N} y_i log\left(\frac{y_i}{m_i}\right)$, where the vector $\overrightarrow{m} = \langle m_1,\ldots,m_N \rangle$ represents the mean distribution of $\overrightarrow{X}$ and $\overrightarrow{Y}$, calculated as: $m_i = \frac{1}{2}(x_i + y_i)$ (where $0 \leqslant x_i, y_i \leqslant 1$ and $\sum_{i=1}^{N} x_i = \sum_{i=1}^{N} y_i = 1$).

The JSD value is *0* for identical distributions, and approaches ln*(2)* ($\sim 0.693$) as the two distributions become more different from one another. We use JSD to calculate the inter-topic distance between each pair of topics, where the distribution dimension $N$ is the number of coded conditions in our vocabulary. In the work presented here, $N = 180$ for the office visit dataset and $N = 250$ for the hospitalization set.

## 3. Experiments and results

We applied the LDA implementation provided via the *R topicmodels* library [32], to the patient-conditions corpus, where each of the resulting topics is a distribution over coded conditions. We experimented with different number of topics, $K$, ranging from *5* to *100*. To avoid the use of poor initial estimates as part of the Gibbs sampling process, we discarded *4,000* samples in the burn-in period – the initial stage of the sampling process in which the Gibbs samples are poor estimates of the posterior. We repeated each experiment five times employing different initial seeds, and calculated an average log-likelihood value. We saved the initial seeds so that the results can be reproduced.

As mentioned in Section 2.2, to determine the optimal number of

**Table 2**

Examples of five of the *13,111* patient-conditions office-visits records. The leftmost column shows patient IDs and the topmost row shows eight of the *180* conditions. Each cell lists the number of times a patient has been diagnosed with the respective condition, during the eight-year period for which data was gathered.

| Conditions / Patient-ID | Hypertension | Type 2 Diabetes Mellitus | Morbid Obesity | Goiter | Vitamin D Deficiency | Anemia | Pain in Limb | Osteoporosis |
|---|---|---|---|---|---|---|---|---|
| **1** | 17 | 17 | 10 | 2 | 2 | 9 | 2 | 2 |
| **2** | 0 | 0 | 0 | 0 | 5 | 0 | 5 | 5 |
| **3** | 8 | 8 | 8 | 0 | 0 | 8 | 0 | 0 |
| **4** | 0 | 0 | 0 | 0 | 20 | 0 | 20 | 20 |
| **5** | 10 | 10 | 10 | 10 | 0 | 10 | 0 | 0 |

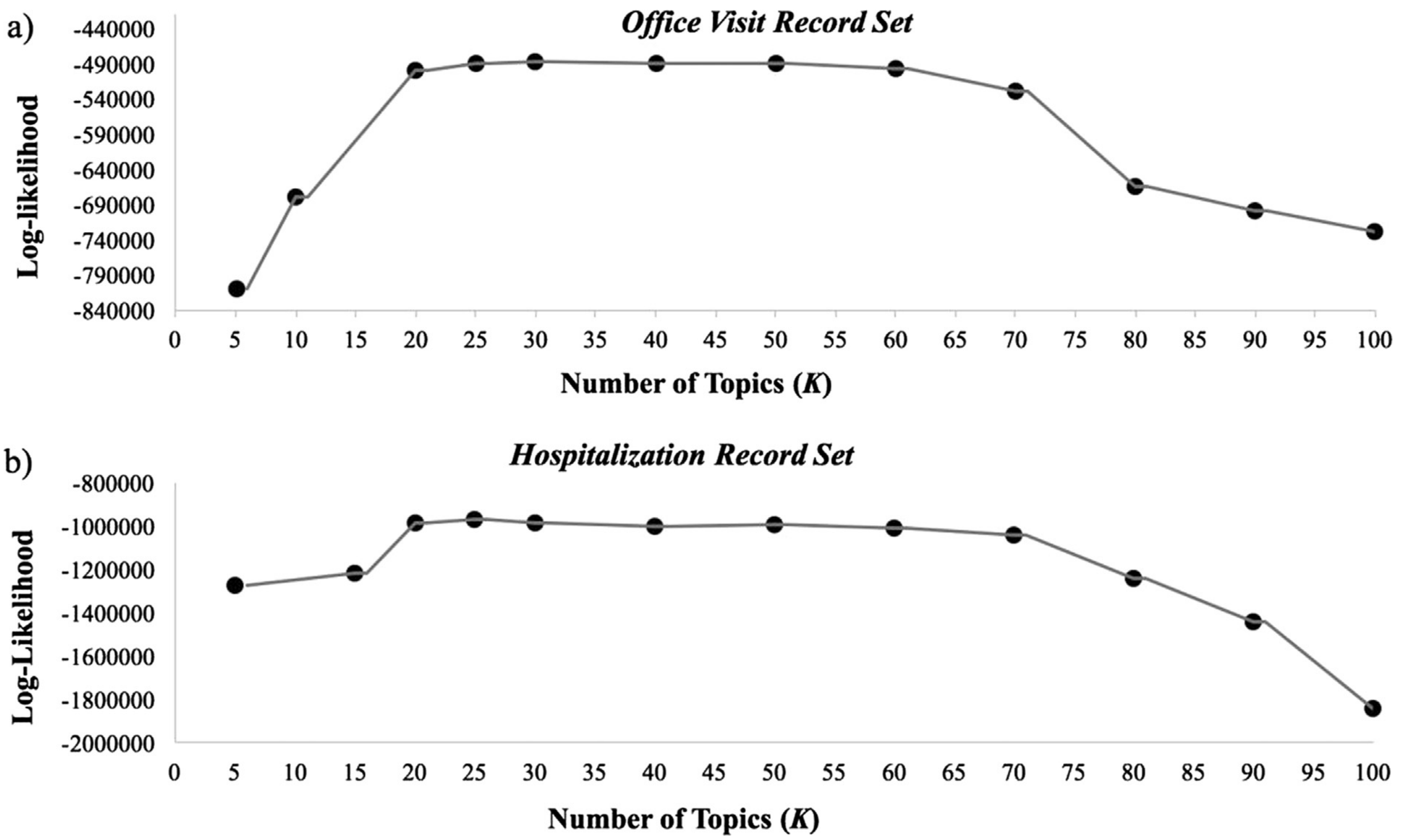

**Fig. 3.** Data log-likelihood as a function of the number of topics, *K*, for the office visit record set (a) and for the hospitalization record set (b). The x-axes correspond to the number of topics (*K*) ranging from *0* to *100*; the y-axes correspond to the data log-likelihood values.

topics, we identified the value *K* that maximizes the data log-likelihood. Fig. 3 shows plots of the data log-likelihood as a function of the number of topics (*K*), for both the office-visit (Fig. 3a) and the hospitalization (Fig. 3b) sets. As demonstrated in the figure, for both record sets, the log-likelihood shows almost no change as *K* ranges between *20* and *30*, while the maximum value log-likelihood is obtained at $K = 30$ for the office visit set and $K = 25$ for the hospitalization set. As *20* topics are easier to describe, visualize and evaluate, we report results obtained for $K = 20$.

Table 3 shows examples of five characteristic topics from the *20* identified by our model from the office visit set. For each of the five displayed topics, we list the conditions that are assigned a probability greater than a threshold value – set here to *0.01*, along with their respective probabilities. The term distributions associated with the other *15* topics show similar characteristics. The results obtained over the hospitalization set are similar (Table 4).

After obtaining the topics, we conducted a thorough evaluation of the performance of our method. First, we surveyed the medical literature by manually searching the *PubMed*, *Google Scholar*, *WebMD* and the *Mayo Clinic* websites [34–47] to verify whether the most probable conditions within each topic are indeed known to co-occur according to medical literature. We then conducted a rigorous quantitative evaluation by assessing the resulting topics in terms of tightness and distinctiveness.

To assess tightness we inspected, for each topic, the number of codes within it that are assigned a probability greater than *0.01* (the threshold value set here). We refer to each list of coded conditions whose probability is greater than *0.01* as the *top-codes* for the topic. As shown in Table 3, four of the five topics (Topics 1–4) are associated with *10* top-codes, while the remaining topic (Topic 5) is associated with *13* top-codes. The cumulative probability mass associated with the top-codes accounts for over *0.9* of the total probability mass, as shown at the bottom row of the table. Our results show that for each topic inferred from the office visits set, only a few codes have a probability above *0.01* (at most *15,* out of *180* codes); moreover, the cumulative probability mass for these *15* codes is above *0.9*.

Fig. 4 shows graphs plotting the probability of each of the *180* codes to be associated with each of four example topics. The *x-axis* corresponds to the *180* codes, while the *y-axis* corresponds to the conditional probabilities of the codes to occur in the respective topic. As shown in the figure, for each of the four topics, only a few codes (less than *15*) are assigned a non-negligible probability of association with the topic, further illustrating tightness of the topics. Likewise, for each topic obtained from the hospitalization set, *25* or fewer codes have a probability higher than the threshold value (*0.01*), and the cumulative probability of these *25* codes is higher than *0.9*.

We also calculated the entropy of each topic and compared the values to the entropy of the uniform distribution (i.e., the maximum possible entropy, $log_2 N$, where *N* is the number of codes in our vocabulary). The entropy for each of the 20 topics obtained from the office visit set ranges from *1.208* to *3.576*; the average entropy value is *2.853* ($\sigma = 0.557$), which is much lower than the entropy of the uniform distribution, ($log_2 N = 7.491$, where $N = 180$). Similarly, for the hospitalization set, the entropy of the 20 topics ranges from *0.849* to *4.907;* the average entropy value is *3.507* ($\sigma = 1.049$); again, much lower than the entropy of the uniform distribution ($log_2 N = 7.966$, where $N = 250$).

To assess *distinctiveness* of the topics, we used two approaches. First, we calculated and compared the corpus-wide versus the topic-specific number of occurrences of each top-code associated with each topic. Fig. 5 shows two example plots depicting both the corpus-wide and the topic-specific abundance of top-codes for topics obtained from the office visit set (*Topics 1* and *4* in Table 3). For each coded condition, the black bar represents the topic-specific number of occurrences, while the combined black-and-grey bar represents the corpus-wide number of occurrences.

We also calculated inter-topic distances between all distinct pairs within the *20* topics inferred from each record set, using Jensen-Shannon divergence [33]. The mean value of the JSD obtained from the office visit set is *0.666* ($\sigma = 0.150$), (with median = *0.691* and minimum = *0.483*), while for the hospitalization set, the mean is *0.637* ($\sigma = 0.053$), with median and min of *0.649* and *0.423*, respectively.

While as noted in Section 2.1 we utilized in the above experiments only *180* (or *250*) codes accounting for *80%* of the cumulative code frequency in the respective dataset, we have conducted additional experiments varying the vocabulary size. Using a vocabulary comprising all 4,000 coded-conditions in the dataset led to very sparse data-matrices and to topics that were neither distinct nor tight. We also used codes accounting for *90%* of the respective cumulative frequencies in each dataset, *590* codes to represent the office visit record set and *616* codes for the hospitalization record set. The topics obtained using these vocabularies were as informative and clinically meaningful – but not as tight – as the topics obtained when the vocabulary was smaller and limited to codes comprising *80%* of the cumulative frequency. We thus only report here the results obtained using the latter smaller vocabulary.

**Table 3**
Examples of five characteristic topics from among the twenty topics identified by our model over the office visit dataset. Each column lists conditions whose association probability with the respective topic is higher than *0.01*, along with the respective probability, $\Pr(C|T_i)$, where $C$ denotes a condition and $T_i$ denotes the $i^{th}$ topic.

| Topic 1 | $Pr(C\|T_1)$ | Topic 2 | $Pr(C\|T_2)$ | Topic 3 | $Pr(C\|T_3)$ | Topic 4 | $Pr(C\|T_4)$ | Topic 5 | $Pr(C\|T_5)$ |
|---|---|---|---|---|---|---|---|---|---|
| Type 2 diabetes mellitus | .369 | Pain in limb | .195 | Hypothyroidism | .400 | Allergic rhinitis | .361 | Atrial fibrillation | .197 |
| Type II diabetes mellitus uncontrolled | .132 | Arthralgia of the lower leg | .166 | Disorder of bone and articular cartilage | .229 | Osteoporosis | .186 | Congestive heart failure | .172 |
| Mixed hyperlipidemia | .099 | Low back pain | .144 | Vitamin D deficiency | .161 | Acute sinusitis | .111 | Arthropathy | .112 |
| Disorder associated with type 2 diabetes mellitus | .088 | Shoulder joint pain | .128 | Anemia | .078 | Benign essential hypertension | .110 | Peripheral venous insufficiency | .085 |
| Neurologic disorder associated with type 2 diabetes mellitus | .074 | Chronic renal failure | .107 | Diaphragmatic hernia | .044 | Female sexual arousal disorder | .076 | Metabolic syndrome | .083 |
| Proteinuria | .064 | Arthralgia of the pelvic region and thigh | .092 | Degenerative joint disease of pelvis | .036 | Chronic sinusitis | .061 | Transient cerebral ischemia | .062 |
| Morbid obesity | .063 | Thoracic radiculitis | .071 | Abnormal findings on diagnostic imaging of breast | .034 | Acute upper respiratory infection | .046 | Cerebral infarction due to thrombosis of cerebral arteries | .061 |
| Benign essential hypertension | .038 | Joint pain | .036 | Goiter | .014 | Disease of liver | .043 | Syncope and collapse | .050 |
| Vitamin D deficiency | .032 | Acute upper respiratory infection | .030 | Benign essential hypertension | .002 | Acute bronchitis | .004 | Conduction disorder of the heart | .049 |
| Diabetic-oculopathy associated with type 2 diabetes mellitus | .029 | Chronic rhinitis | .029 | Conduction disorder of the heart | .001 | Impacted cerumen | .002 | Cellulitis-abscess of lower leg | .043 |
| | | | | | | | | Type 1 diabetes Mellitus | 0.038 |
| | | | | | | | | Non-toxic multinodular goiter | 0.034 |
| | | | | | | | | Gastro intestinal hemorrhage | 0.034 |
| *CUMULATIVE PROBABILITY* | *.989* | | *.998* | | *1.00* | | *1.00* | | *.997* |

Last, as also mentioned earlier, in addition to employing bag-of-codes to represent each patient-file, we have experimented using *binary* (*0/1*) and *tf-idf* representations. The former representation does not account for code abundance, while the latter penalizes code abundance if a code occurs in many patient records. While the complete results are not shown here due to space limits, the topics resulting from the 0/1 representation are not as distinct as the ones obtained using bag-of-codes, although topics inferred using each of the two representations capture many similar condition-association patterns. Moreover, since *tf-idf* penalizes codes that frequently appear within the corpus, this representation loses important information, as a persistently recurring condition (i.e. a code with a high count) is often indicative of a stronger association with other conditions. Hence, topics formed under the tf-idf representation do not capture some of the associations between recurring conditions (e.g. hypertension), and other conditons that are manifested in the disease. We have thus focused our report on the bag-of-codes representation.

## 4. Discussion

As demonstrated by Tables 3 and 4, the topics obtained from both the office visit and the hospitalization datasets indeed reveal patterns of commonly co-occurring conditions. As mentioned in Section 1, while similar disease themes are found in topics inferred from the hospitalization and from the office visit sets (e.g. Topic *5* in Table 3, and Topics *4* and *5* in Table 4 are all related to *Heart* disease), disease profiled by topics emerging from hospitalization records reflect a higher level of severity than the topics stemming from office visits. This difference in co-occurrence patterns matches the reality in which hospitalized patients often show more severe manifestations of the disease than non-hospitalized patients. Fig. 1 illustrates this point, where the twenty most frequent codes associated with the hospitalization set denote more severe conditions than those associated with the office visit set. For example, *congestive heart failure*, *atrial fibrillation*, and *acute renal failure syndrome*, are among the most frequent condition codes in the hospitalization list (Fig. 1b) but are not among the most frequent codes in the office visit set (Fig. 1a).

To validate that the condition-association patterns exposed through topic models are meaningful, we verify that the most probable conditions within each topic have indeed been reported to co-occur in the medical literature. For instance, many of the conditions grouped together in the leftmost column of Table 3, (*Topic 1*), are clearly related to Diabetes – one of the most frequent causes of decline in kidney function [34–37]. Likewise, most of the conditions grouped in *Topic 2* (see Table 3) are related to Limb- or Joint-pain, including Chronic Renal-failure, since Metabolic Bone Disease is a frequent complication of advanced kidney disease [3,38]. Similar relationships characterize each of the other topics [34–47].

To further assess the clinical validity of our topics, we selected *10* topics, examining two randomly selected high probability codes from each. We surveyed the medical literature to verify whether these coded conditions are known to co-occur [34–47]. Table 5 shows for each pair of conditions, the PubMed identifier of the paper that establishes the medical connection between them. Most of the condition pairs are well known according to the medical literature, supporting the hypothesis that the model adequately uncovers true associations among coded conditions.

**Table 4**
Examples of five characteristic topics from among the twenty topics identified by our model over the *hospitalization* dataset. Each column lists description of the top-codes whose association probability with the respective topic is higher than *0.01*, along with the respective probability, $Pr(C|T_i)$, where $C$ denotes a coded-condition and $T_i$ denotes the $i^{th}$ topic.

| Topic 1 | $Pr(C\|T_1)$ | Topic 2 | $Pr(C\|T_2)$ | Topic 3 | $Pr(C\|T_3)$ | Topic 4 | $Pr(C\|T_4)$ | Topic 5 | $Pr(C\|T_5)$ |
|---|---|---|---|---|---|---|---|---|---|
| Hypertensive Renal Disease with Renal Failure | .211 | Chronic Obstructive Lung Disease | .220 | Gastroesophageal Reflux Disease | .325 | Atrial Fibrillation | .444 | Congestive Heart Failure | .325 |
| End Stage Renal Disease | .171 | Tobacco User | .192 | Osteoarthritis | .139 | Patient with Cardiac Pacemaker | .097 | Cardiomyopathy | .110 |
| Anemia of Chronic Renal Failure | .149 | Dyspnea | .058 | Osteoporosis | .089 | Benign Prostatic Hypertrophy without Outflow Obstruction | .078 | Automatic implantable cardiac defibrillator | .094 |
| Dialysis Finding | .138 | Pneumonia | .052 | Rheumatoid Arthritis | .067 | Hyperlipidemia | .068 | Chronic ischemic heart disease | .078 |
| Hyper *para*-thyroidism due to Renal Insufficiency | .047 | Acute Exacerbation of Chronic Obstructive Airways Disease | .046 | Anemia | .059 | Allergic Reaction | .037 | Chronic pulmonary heart disease | .072 |
| Renal Osteodystrophy | .039 | Hypoxemia | .040 | Glaucoma | .047 | Atrial Flutter | .034 | Mitral Valve Disorder | .055 |
| Renal Disorder Associated with type 2 Diabetes Mellitus | .028 | Chronic Pulmonary Heart Disease | .039 | Hearing Loss | .042 | Blood Coagulation Disorder | .033 | Acute Chronic Systolic Heart Failure | .052 |
| Nephritis | .027 | Hyperlipidemia | .034 | Systemic Lupus Erythematosus | .031 | Sinus Node Dysfunction | .032 | Chronic Systolic Heart Failure | .049 |
| Hyperkalemia | 0.026 | Chronic Asthmatic Bronchitis | .026 | Vitamin D Deficiency | .023 | Conduction Disorder of the Heart | .021 | Left Bundle Branch Block | .029 |
| Hypervolemia | 0.017 | Acute Respiratory Failure | .024 | Diaphragmatic Hernia | .022 | Right Bundle Branch Block | .017 | Dyspnea | .028 |
| CKD 4 | 0.017 | Chronic Diastolic Heart Failure | .022 | Localized Osteoarthrosis | .020 | Pleural Effusion | .013 | Paroxysmal Ventricular Tachycardia | .024 |
| Complication of Internal Anastomosis | 0.016 | Acute Chronic Diastolic Heart Failure | .018 | Dehydration | .018 | Rheumatic Disease of Tricuspid Valve | .013 | Hypertensive Heart and Renal Disease with Congestive Heart Failure | .021 |
| Complication of Implant | 0.015 | Post Inflammatory Pulmonary Fibrosis | .018 | Constipation | .018 | Chronic Diastolic Heart Failure | .010 | Rheumatic Disease of Tricuspid Valve | .016 |
| Disorder of Phosphorus Metabolism | 0.015 | Congestive Heart Failure | .018 | Disorder of Bone and Articular Cartilage | .010 | Hypoxemia | .010 | Low Blood Pressure | .010 |
| Disorder of Bone and Articular Cartilage | 0.015 | Acute Chronic Respiratory Failure | .015 | | | | | | |
| CKD 5 | 0.014 | Constipation<br>… | .014 | | | | | | |
| ***CUMULATIVE PROBABILITY*** | .945 | | .903 | | .924 | | .907 | | .963 |

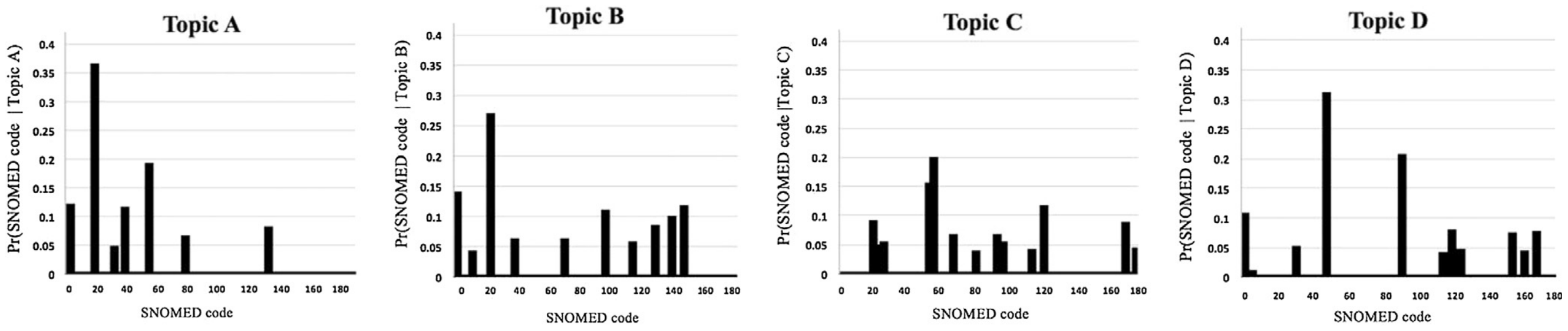

**Fig. 4.** The probability distribution of SNOMED codes in four example topics denoted A-D identified by our model when applied to the office visit set. The x-axes correspond to the *180* codes; the y-axes correspond to the conditional probability of each code to occur in the respective topic.

Notably, a few of the coded condition pairs grouped together in our topics are not known to be directly associated with each other. For instance, no direct association is reported between *Allergic Rhinitis* and *Osteoporosis*, two conditions grouped under *Topic 4*. However, a recent study has shown that treating Allergic Rhinitis with depot-steroid injections increases the risk of Osteoporosis [39], which supports the high correlation we find between the two conditions. Similarly, *Renal failure* and *Bone- and Joint-pain* are not reported as associated conditions; however, Renal failure causes *Osteodystrophy*, which in turn causes Bone- and joint-pain [3], explaining the significant association exposed among the conditions in our results. The identification of two indirect medical associations, suggests that our approach is likely to expose yet unnoticed associations among medical conditions when applied to additional datasets.

Our quantitative evaluation of the model performance shows that the obtained topics are indeed tight and distinct. A topic that can be specified by a small number of coded conditions (tightness), is expected to capture meaningful and specific associations, since the codes characterizing the topic are highly correlated with each other.

Fig. 4 shows that each plot contains only a small number of peaks (typically *10*), indicating that no more than 10 coded conditions account for most of the topic's probability mass, demonstrating the tightness of our resulting topics. Thus, *10* conditions are typically sufficient for characterizing each topic inferred from the office visit record set, although a few topics are characterized by *11–15* coded conditions (e.g. Topic 5, rightmost column of Table 3).

The probability plots for the other *16* topics show very similar trends, and are not shown here due to limited space. Moreover, the significantly lower entropy of the topic-distributions compared to the entropy of the uniform distribution (the max-entropy distribution) indicates that the topics obtained by our model bear high information contents. The information-based assessment is extended as part of the evaluation of distinctiveness, using the Jensen-Shannon divergence as discussed below.

Topic distinctiveness is established by showing that many of the codes appearing in a topic's top-codes list are predominantly associated with a specific topic. That is, a top-code corpus-wide abundance is close in count to its topic-specific abundance for one particular topic. For instance, Fig. 5 shows that the majority of the top-codes associated with *Topic 1* (eight of ten) and with *Topic 4* (six of ten) are all-black, that is, the topic-specific count is the same as the corpus-wide count for these codes. We likewise inspected the other 18 topics (not shown here), observing a similar trend, indicating that the topics capture distict patterns of co-occurrences. While most codes are typically associated with a single topic, a few codes, such as *benign essential hypertension*, *diabetes* and *vitamin D deficiency*, are strongly associated with multiple topics as these conditions co-occur with several different groups of conditions.

The large inter-topic distance between each pair of topics, indicated by high JSD values, further illustrates the distinctiveness of the resulting topics. As discussed earlier, JSD values range from *0* to ln*(2)* (~0.693), where *0* indicates identical distributions, and ln*(2)* indicates orthogonal distributions. The higher the divergence value between two topics, the more distinct they are from one another. The high mean and median values (close to the upper bound of ln(*2*)) of the inter-topic distances we have obtained indicate that almost all our topic are highly distinct.

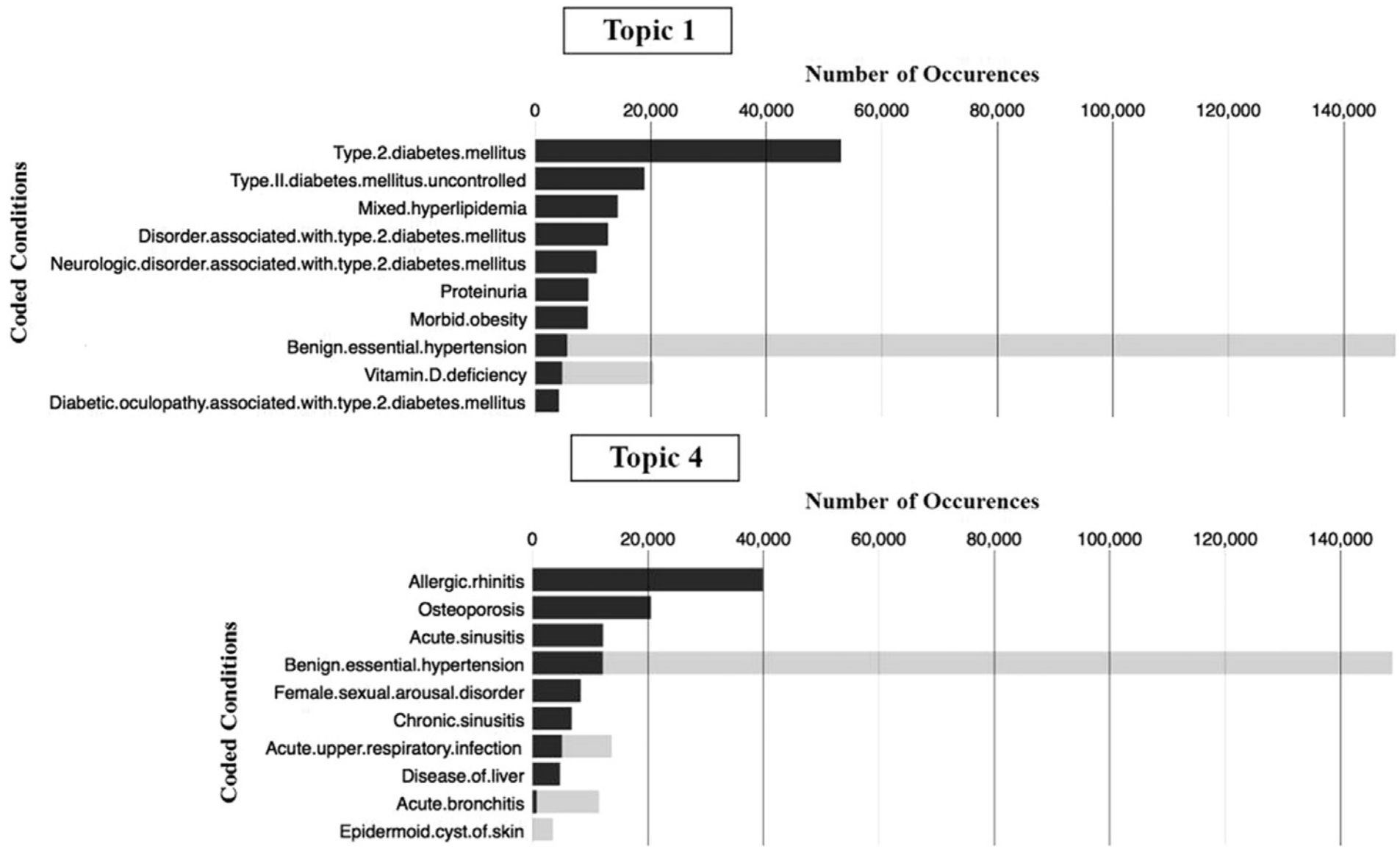

**Fig. 5.** Topic-specific and corpus-wide number of occurrences of the top-codes associated with Topics 1 and 4, obtained from the office visit set. Black bars represent topic-specific number of occurrences for each condition. Combined black-and-grey bars represent the corpus-wide number of occurrences for each condition.

**Table 5**
Medical relevance of associations between condition-pairs grouped in topics obtained from the office visits set. Column 1 indicates the topic number; Columns 2 and 3 list two randomly selected conditions that have a high probability to be associated with the topic; Column 4 provides the PUMED identifier and the reference number of the paper establishing the medical conditions as co-occurring.

| Topic | Condition 1 | Condition 2 | Supporting literature |
|---|---|---|---|
| 1 | Proteinuria | Type 2 diabetes mellitus | 7050509 [40] |
| 2 | Chronic renal failure | Arthralgia of the lower leg | 17251386 [38] |
| 3 | Atrial fibrillation | Congestive heart failure | 20347787 [41] |
| 4 | Allergic rhinitis | Osteoporosis | 24090789 [39] |
| 6 | Hyperlipidemia | Essential hypertension | 1927888 [42] |
| 7 | Gastroesophageal reflux | Insomnia | 20535322 [43] |
| 8 | Hypothyroidism | Disorder of bone and articular cartilage | 24783033 [44] |
| 9 | Coronary arteriosclerosis | Disorder of kidney and/or ureter | 24527682 [45] |
| 10 | Anemia | Gout | 22906142 [46] |
| 11 | Chronic obstructive lung disease | Tobacco dependence syndrome | 17040932 [47] |

The topics obtained from the hospitalization records are also clinically-relevant, tight and distinct. As with the office visit records, we surveyed the medical literature and found that the coded conditions grouped together as topics, inferred from the hospitalization dataset, have been reported to co-occur. Moreover, the observation that only a few coded conditions (25 or fewer, out of *250*) characterize each topic, indicates tightness, while the high JSD values of the topics indicate distinctiveness. Furthermore, the low entropy of the topic distributions (compared to that of a uniform distribution), indicates that the topics are indeed informative. The ability of our method to generate clinically-relevant, tight and distinct topics from both datasets, demonstrates its effectiveness in identifying patterns of co-occurring medical conditions.

## 5. Conclusion

We have employed topic modeling over disease codes using LDA, obtaining probabilistic topics that highlight characteristic patterns of co-occurring medical conditions among patients who have kidney disease. Our results indicate that most coded conditions grouped together within a topic, are indeed reported to co-occur in the medical literature. Our results also uncover several associations among conditions that were hitherto not reported as co-occurring. We quantitatively evaluated the performance of our method and have shown that the topics identified from two different datasets are tight and distinct.

Our approach can also be helpful as a basis for a recommender system, suggesting to the practitioner, conditions that are likely to co-occur with a patient's current diagnoses. Given a diagnosis code, the topic with which the code is most strongly associated can be identified. Accordingly, the list of the other codes associated with the topic can be shown to the physician, as conditions to be checked for. For instance, if a healthcare provider seeing a patient with decreased kidney function enters Anemia as a diagnosis, the system will prompt checking for Goiter, Hypothyroidism and Hypertension, since these conditions are all strongly associated with the same topic, thus highly likely to co-occur. Last, as conditions associated with other diseases are coded in a similar way within EHRs, we expect that our proposed method can be used to address similar research questions for diseases beyond kidney dysfunction.

## Acknowledgments

This research was partially supported by the NIGMS IDeA grants U54-GM104941 and P20 GM103446, and by NSF IIS EAGER grant #1650851. We thank James T. Laughery and Sarahfaye Dolman for their major role in building the dataset.